\documentclass{article}

\PassOptionsToPackage{numbers}{natbib}

 \usepackage[preprint]{nips_2018}

\usepackage[utf8]{inputenc} 
\usepackage[T1]{fontenc}    
\usepackage{hyperref}       
\usepackage{url}            
\usepackage{booktabs}       
\usepackage{amsfonts}       
\usepackage{nicefrac}       
\usepackage{microtype}      
\usepackage{graphicx}
\usepackage{float}

\title{Embedding of FRPN in CNN architecture}

\author{
  Alberto Rossi \\
  Department of Information Engineering\\
  University of Florence\\
  Florence, FI 50139 \\
  \texttt{alberto.rossi@unifi.it} \\
  \And
  Markus Hagenbuchner \\
  School of Computing and Information Technology\\
  University of Wollongong \\
  Northfields Avenue, 2522 Wollongong - Australia \\
  \texttt{markus@uow.edu.au} \\
  \And
  Franco Scarselli \\
  Department of Information Engineering \\ 
  University of Siena \\
  Via Roma 56, 53100 Siena - Italy \\
  \texttt{franco@diism.unisi.it} \\
  \And
  Ah Chung Tsoi \\
  School of Computing and Information Technology \\
  University of Wollongong \\
  Northfields Avenue, 2522 Wollongong - Australia \\
  \texttt{act@uow.edu.au} \\
}

\begin{document}

\maketitle

\begin{abstract}
This paper extends the fully recursive perceptron network (FRPN) model for vectorial inputs to include deep convolutional neural networks (CNNs) which can accept multi-dimensional inputs. A FRPN consists of a recursive layer, which, given a fixed input, iteratively computes an equilibrium state. The unfolding realized with this kind of iterative mechanism allows to simulate a deep neural network with any number of layers. The extension of the FRPN to CNN results in an architecture, which we call convolutional-FRPN (C-FRPN), where the convolutional layers are recursive. The method is evaluated on several image classification benchmarks. It is shown that the C-FRPN consistently outperforms standard CNNs having the same number of parameters. The gap in performance is particularly large for small networks, showing that the C-FRPN is a very powerful architecture, since it allows to obtain equivalent performance with fewer parameters when compared with deep CNNs.
\end{abstract}

\vspace*{-1mm}
\section{Introduction}
\vspace*{-1mm}

Convolutional Neural Networks (CNNs) emerged in the last decade as a leading method to process images~\cite{lecun1998gradient,electronics8030292}. Modern CNNs have deep and wide-width architectures~\cite{electronics8030292}. 
One of the first  CNNs 
\cite{lecun1998gradient} was composed of five layers, two of which were convolutional.
The depth of the models in subsequent years showed a continuously increasing trend. The AlexNet  \cite{krizhevsky2012imagenet} and the ZFNet \cite{zeiler2014visualizing} architectures, which won the object recognition task at ILSRVC in 2012 and 2013, respectively, have  8 layers. 
GoogLeNet, 
the winning model at ILSRVC~\cite{szegedy2015going} in 2014, is a CNN composed of 22 layers.
A major advance in performance was obtained in the following year by ResNet~\cite{he2016deep}, based on a highway network architecture, consisting of 152 layers.
Such a relationship between the increase in performance level and the depth of the network was observed in many other image related tasks, e.g., object detection, semantic segmentation, image generation~\cite{redmon2016you,chen2017deeplab,isola2017image,andreini2019two}.

Despite the important role played by the depth of the architectures, such a meta--parameter is still chosen by heuristics and by a trial and error process. Moreover, the depth, once determined, remains fixed during the learning and application phase and for each input exemplar class. In order to overcome these  limitations, in this paper, we extend the fully recursive perceptron network (FRPN)~\cite{hagenbuchner2017fully} to include CNNs. The FRPN is an architecture in which all the hidden neurons are fully connected with a weight parameters~\cite{hagenbuchner2017fully}. Intuitively, the hidden layer of a FRPN implements a recursive layer, that, given a fixed input, iteratively computes a hidden state until the state itself converges~\cite{hagenbuchner2017fully}.

The extended model, which we call a convolutional-FRPN (C-FRPN), is characterized by convolutional layers having  weighted parameter feedback connections. In those layers, the convolutions are applied on the concatenation of the output of the convolutions at the previous iteration with the static input coming from the previous layer, realizing a kind of state iteration on feature maps.
The computation evolves until the feature maps converge. In this way, the unfolding of a C-FRPN is a deep network, where the number of layers is not fixed, but adapted automatically to suit a given task or to suit a particular input sample.

The proposed architecture is related to recurrent convolutional neural networks (RCNNs)~\cite{liang2015recurrent}, which use recurrent layers,
and reported very good results on several image classification tasks. However, in RCNNs, the number of iterations is predefined and thus the unfolded architecture is of a fixed depth, and requires a manually defined hyper-parameter for each recursive layer, whereas in C-FRPN such a parameter is learned.

This paper studies the effectiveness of the proposed method via experimentation on standard natural image benchmark datasets (CIFAR-10 and SVHN), and on a real-world problem on melanoma prediction (the ISIC dataset). The experiments revealed that C-FRPNs outperform CNNs having the same number of weight parameters. The difference is particularly evident for smaller architectures where the number of weights is small. Those results suggest that the C-FRPN model is more powerful in terms of approximation capability than the standard CNN with the same number of parameters. Such a capability is easy to  be exploited if a sufficient number of training examples is available, and allows to solve a problem with applications where computational resources are limited.

The rest of the paper is organized as follows: Section~\ref{sec:FRPN} recalls the FRPN model, Section~\ref{sec:C-FRPN} introduces the C--FRPN architecture, Section~\ref{sec:Experiments} describes the used datasets, the experimental settings and the obtained results. Finally, Section \ref{Sec:Conclusion} offers concluding remarks. 

\vspace*{-1mm}
\section{The FRPN model}\label{sec:FRPN}
\vspace*{-1mm}
A FRPN is a neural network with an input layer, a single hidden layer and an output layer. The peculiarity of a FRPN is the hidden layer, the neurons of which are fully connected with the other, including with themselves, and to the inputs via  links with unknown weights. Formally,  if  $u = [u_1,\ldots, u_m]$ is an input vector and $x(t) = [x_1(t), \ldots, x_n(t)]$ is the output of the hidden layer at time $t$, then 
\vspace*{-1mm}
\begin{equation}
    x_i(t) = f(\sum_{j=1}^m \alpha_{ij}u_j + \sum_{k=1}^n \beta_{ik}x_k(t-1) + b_i)
\label{eq:state}
\end{equation}
\vspace*{-1mm}
where $b_i$ are bias weights, $\alpha_{ij}$ are the input to hidden connection weights, and $\beta_{ik}$ are the hidden to hidden connection weights, for  $i = 1, 2,..., n$, $j = 1, 2,..., m$ and $k = 1, 2,..., n$. 


Equation~(\ref{eq:state}) defines a dynamic System. In order to calculate the outputs of the hidden neurons, the computation in Equation~(\ref{eq:state}) is iterated until the state $ x$ converges to a stable point or until a maximum number of iterations is reached. In practice, this allows to unfold the network for a number of iterations that is not pre-defined, but it depends on the ``richness'' of the input\footnote{A ``rich'' input excites the latent modes in the system and would characterize the behaviour of the underlying system. This is an essential assumption in any system identification study.}. Moreover, the shared weights in the unfolding network give an opportunity to have a deep network architecture, requiring a small number of weights. Indeed, in \cite{hagenbuchner2017fully} it is formally shown that any deep multi-layer feedforward neural network can be simulated by an FRPN, and predicts advantages of the latter model (FRPN) over the former (deep CNN) in terms of approximation capabilities.

\vspace*{-1mm}
\section{C-FRPN}\label{sec:C-FRPN}
\vspace*{-1mm}
\begin{figure}[t]
    \centering
    \includegraphics[width=\textwidth]{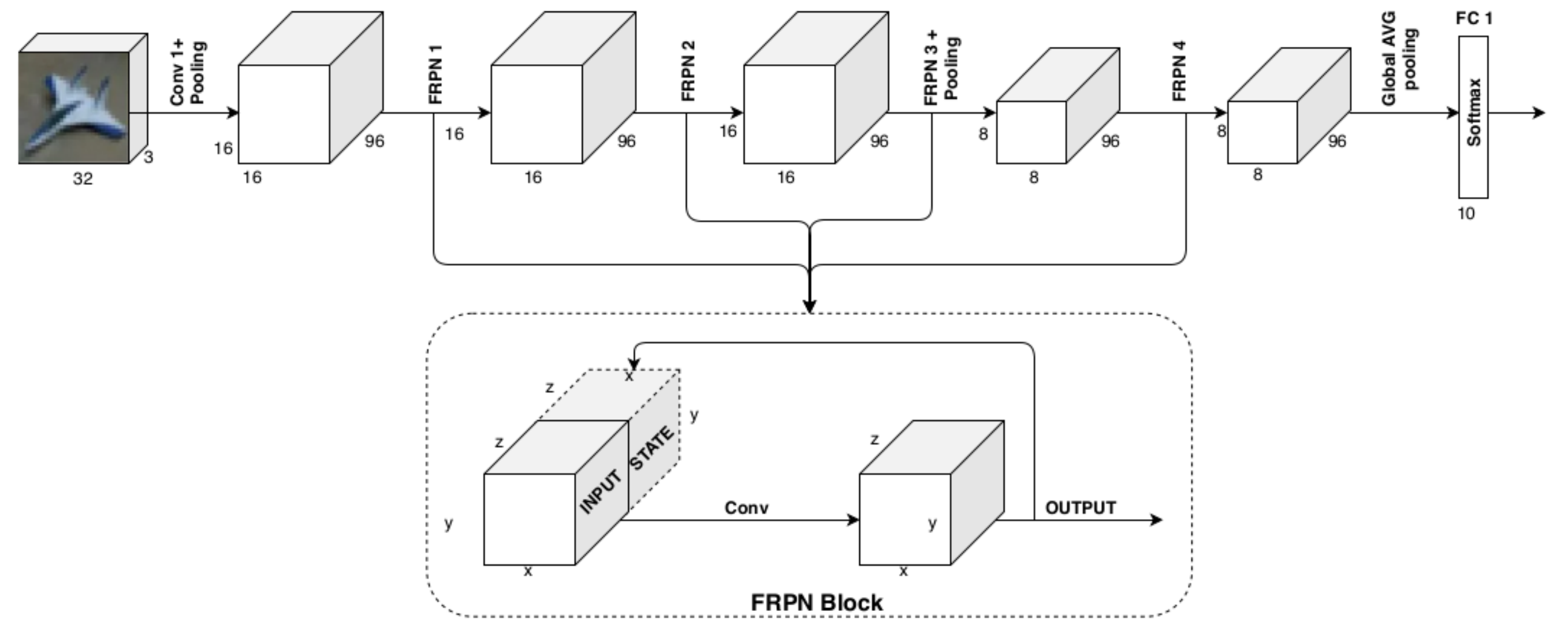}
\vspace*{-5mm}
\caption{The C-FRPN architecture. On top the whole network constituted by 4 C-FRPN layers. On bottom,   a C-FRPN layer.}
    \label{CFRPN}
\end{figure}

A C-FRPN architecture (top of Fig.~\ref{CFRPN}) consists of a CNN in which the convolutional layers are  C-FRPN layers. Each C-FRPN layer computes a set of feature maps taking in input a stack of the feature maps provided by the previous layer and the feature maps of the same layer at the previous time step (bottom of Fig. \ref{CFRPN}). C-FRPN layers behave as FRPNs, where the feature maps, which correspond to  the state, are iteratively computed until they reach a stable point. As in conventional CNN, a C-FRPN layer may be followed by dropout and/or batch normalization layers. Moreover, the architecture may include at the end some full connected layers as is illustrated in Fig.~\ref{CFRPN}.

This architecture is similar to the RCNN proposed in~\cite{liang2015recurrent}. The RCNN consists of $4$ layers
 and each of these unfold exactly $3$ times. At each iteration each layer takes always the same input concatenated with the output evaluated at the previous time step. The RCNN produced  interesting results on three different datasets, proving the benefits of this kind of network.

The first novelty of our approach regards the possibility for each layer to evolve until the state converges to a stable point. This means that the recursive layers are not constrained to unfold for a fixed number of times and such a hyper-parameter has not to be manually set. Furthermore, we obtain a variable depth network: to the best of our knowledge this is the first time that such an architecture is proposed for CNNs.
Another improvement resides in the experimentation carried out that allows to compare the performance of C-FRPN and CNN and disclose interesting differences in their behaviour. 

\vspace*{-1mm}
\section{Experimentation}\label{sec:Experiments}
\vspace*{-1mm}

Experiments have been carried out on three datasets: CIFAR-10~\cite{krizhevsky2009learning}, SVHN~\cite{netzer2011reading} and ISIC~\cite{codella2018skin}. We used a C-FRPN having the overall structure as the one deployed in \cite{liang2015recurrent}. The C-FRPN was compared with a baseline CNN having the same number of layers and the same number of unknown parameters. The architecture included 4 convolutional layers with ReLu activation function and the same number of feature maps on all layers: 6 different experiments have been run, where the feature maps were  $[135,120,104,85,42,21]$ for the baseline CNN and $[96,85,74,60,30,15]$ for the C-FRPN, respectively. The kernel size of the first convolution is $5\times5$ with stride $1$, while for the remaining convolutional layers are $3\times3$ with stride $1$. Each C-FRPN layer was followed by a pooling layer of size $3\times3$  and stride $2$. Moreover, we added a local response normalization operation after each iteration of the C-FRPN layer and a dropout layer with a forget rate of $0.5$ after each convolution except the last one. The state convergence was evaluated by means of the Euclidean distance between the current state and the previous one: the state was considered converged when the distance was smaller than $0.1$ or a maximum number of 8 iterations was reached\footnote{A maximum number of iterations has to be set both for computational reasons and because the convergence cannot be guaranteed without further considerations.}. The image augmentation method in \cite{liang2015recurrent} was deployed for CIFAR-10 and SVHN, while for ISIC, we used random horizontal and vertical flips, and random rotations. The Adam optimizer  was used with  learning rate $1\times10^{-4}$, weight decay $5\times10^{-4}$, and, batch size $128$ for CIFAR-10 and SVHN, and 24 for ISIC. The experiments were repeated 5 times by using different random initial conditions.

\begin{figure}[!t]
    \centering
    \includegraphics[width=\textwidth]{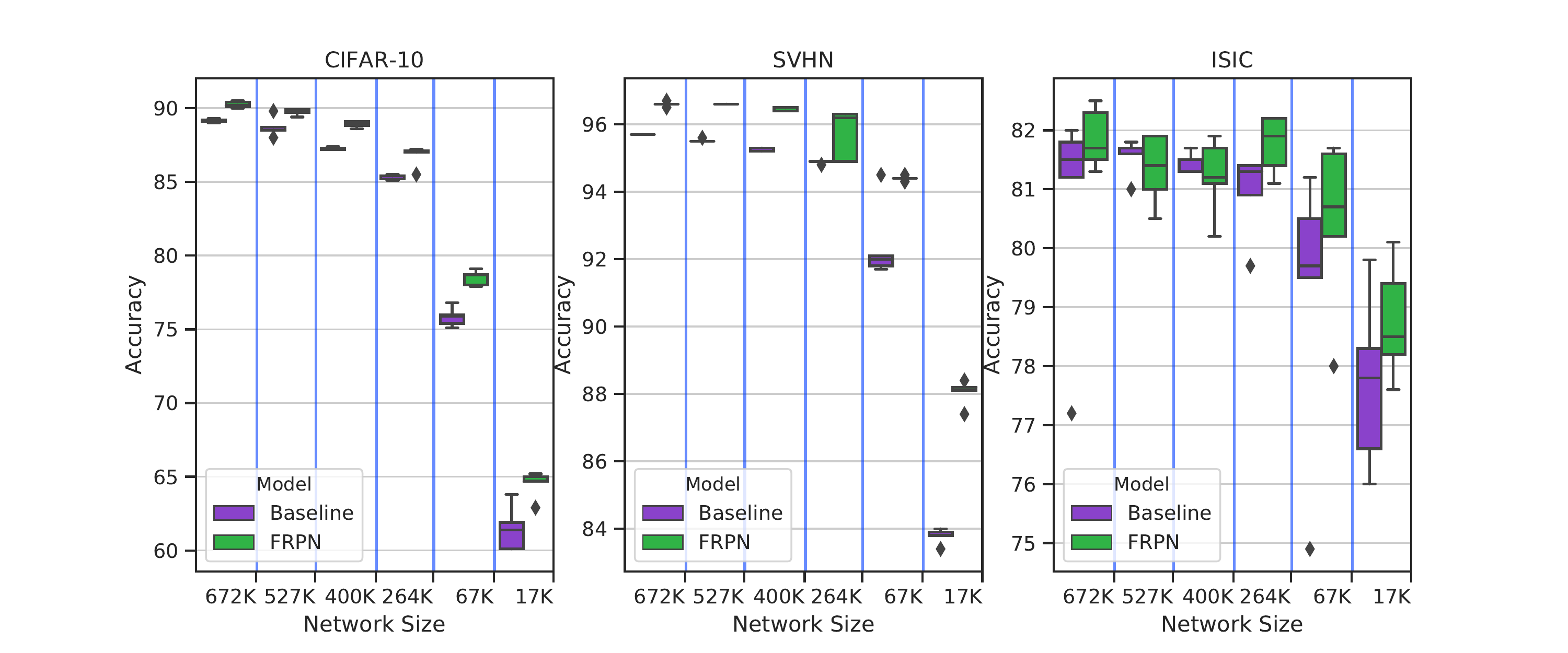}
    \vspace*{-4mm}
    \caption{A box plot of accuracies achieved by C-FRPN and the baseline.}\label{fig:boxPlot}
    \vspace*{-4mm}
\end{figure}
Fig.~\ref{fig:boxPlot} summarizes the results. The figure reveals  that the C-FRPN achieves a higher accuracy when compared with a baseline having the same architecture and the same number of parameters. The difference is more evident for smaller networks. 
\begin{figure}[!t]
    \centering
    \includegraphics[width=\textwidth]{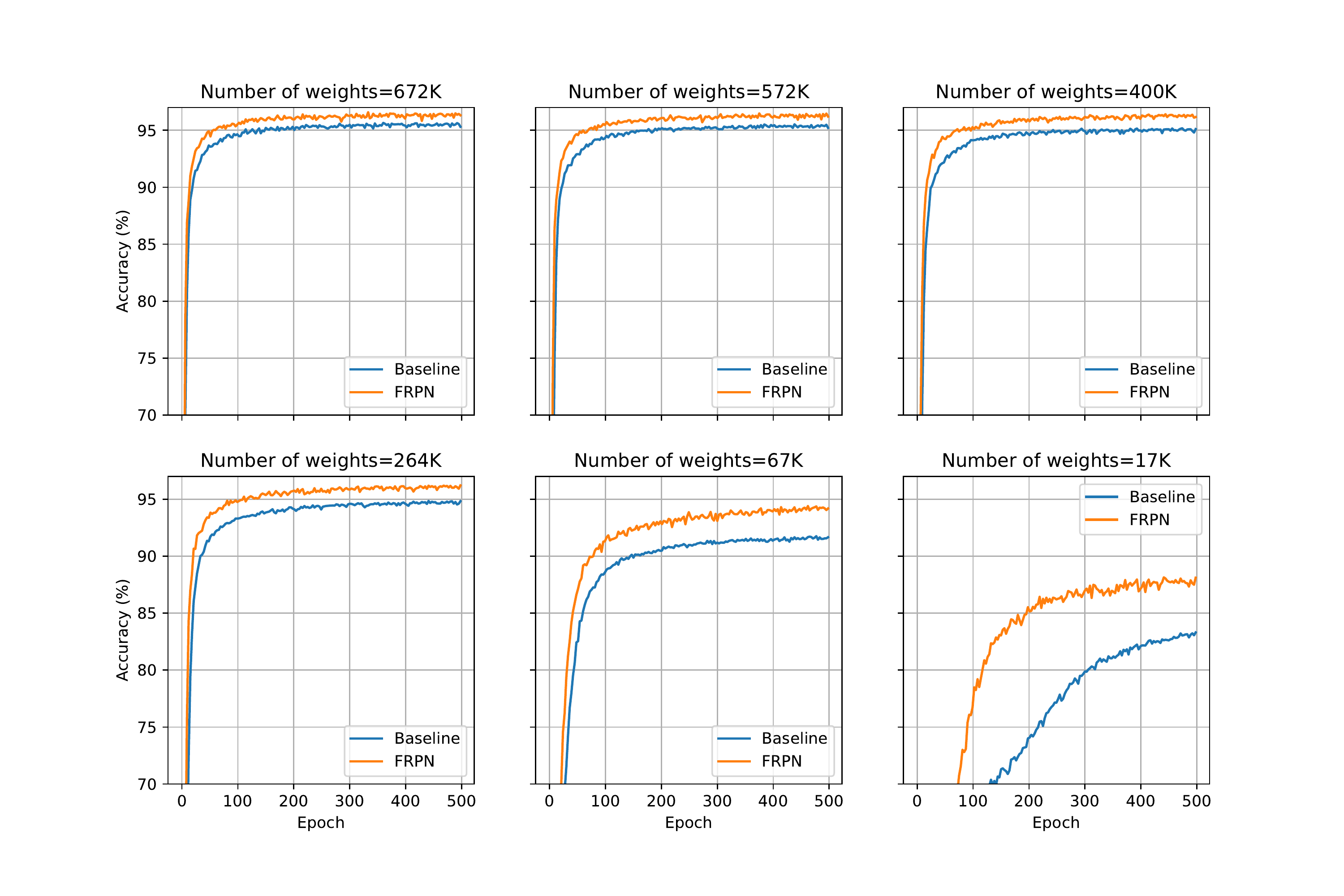}
    \vspace*{-10mm}
    \caption{The validation performance of C-FRPN and the baseline CNN during learning the SVHN data.}
    \label{fig:Generalization}
\end{figure}
Fig.~\ref{fig:Generalization} analyses the validation performance of a single run on the SVHN dataset. It can be observed that the C-FRPN performance is better throughout a training session and that such improvement is more evident for smaller networks. The results suggest that the C-FRPN model is more powerful in terms of approximation capability than the standard CNN. Such a capability can facilitate applications with constraints in computational power and corresponding memory load.


\vspace*{-1mm}
\section{Conclusion}\label{Sec:Conclusion}
\vspace*{-1mm}

This paper proposed the C-FRPN model, which can be considered a generalization of both the FRPN~\cite{hagenbuchner2017fully} and the RCNN~\cite{liang2015recurrent} architectures. The
C-FRPN model realizes variable depth CNNs. An experimental evaluation revealed consistent advantages of the novel architecture, particularly for a small number of parameters. As matters of future research, a more extensive set of experiments and a deeper study of the role played by each C-FRPN layer in the architecture would be most beneficial in understanding the capabilities of this novel architecture.   
\bibliographystyle{plain}
\bibliography{mybibliography}

\end{document}